\documentclass[letterpaper, 10 pt, conference]{ieeeconf}  

\IEEEoverridecommandlockouts                              

\usepackage{amsmath} 
\usepackage{graphicx}
\usepackage{gensymb}
\usepackage{multirow}
\usepackage{subcaption}
\usepackage{algorithm,algorithmic}
\usepackage{pdfpages}
\usepackage{mars}
\usepackage[obeyspaces]{url}

\title{\LARGE \bf
Improving CNN-based Planar Object Detection\\ with Geometric Prior Knowledge}

\author{Jianxiong Cai$^{*}$$^{1}$, Jiawei Hou$^{*}$$^{2}$, Yiren Lu$^{1}$, Hongyu Chen$^{1}$, Laurent Kneip$^{1}$ and S\"oren Schwertfeger$^{1}$ 
\thanks{$^{*}$Both authors are first author and denote equal contribution. 
$^{1}$These authors are with the School of Information Science and Technology, ShanghaiTech University, Shanghai, China.
{\tt\small \{caijx, luyr , chenhy3, lkneip, soerensch\}@shanghaitech.edu.cn} 
$^{2}$Jiawei Hou is with the School of Information Science and Technology, ShanghaiTech University, Shanghai 201210, China, and also with the University of Chinese Academy of Sciences, Beijing, China, and also with the Shanghai Institute of Microsystem and Information Technology, Chinese Academy of Sciences, Shanghai, China.
{\tt\small houjw@shanghaitech.edu.cn} }
}

\begin{document}
%


 \marsPublishedIn{Accepted for:} 		

\marsVenue{IEEE International Symposium on Safety, Security, and Rescue Robotics (SSRR) 2020}

\marsYear{2020}

\marsPlainAutors{Jianxiong Cai$^{*}$, Jiawei Hou$^{*}$, Yiren Lu, Hongyu Chen, Laurent Kneip and S\"oren Schwertfeger}


\marsMakeCitation{Improving CNN-based Planar Object Detection with Geometric Prior Knowledge}{IEEE Press}

\marsDOI{}

\marsIEEE{}


\makeMARStitle

%
%

\maketitle
\thispagestyle{empty}
\pagestyle{empty}


\newcommand{\marky}[1]{#1}

\begin{abstract}
\par
In this paper, we focus on the question: how might mobile robots take advantage of affordable RGB-D sensors for object detection? Although current CNN-based object detectors have achieved impressive results, there are three main drawbacks for practical usage on mobile robots: 1) It is hard and time-consuming to collect and annotate large-scale training sets. 2) It usually needs a long training time. 3) CNN-based object detection shows significant weakness in predicting location.
We propose an improved method for the detection of planar objects, 
which rectifies images with geometric information to compensate for the perspective distortion before feeding it to the CNN detector module, typically a CNN-based detector like YOLO or MASK RCNN. By dealing with the perspective distortion in advance, we eliminate the need for the CNN detector to learn that. Experiments show that this approach significantly boosts the detection performance.  Besides, it effectively reduces the number of training images required. 
In addition to the novel detection framework proposed, we also release an RGB-D dataset and source code for hazmat sign detection. 
To the best of our knowledge, this is the first work of image rectification for CNN-based object detection, and the dataset is the first public-available hazmat sign detection dataset with RGB-D sensors.

\par

\end{abstract}





\section{INTRODUCTION}

\begin{figure}[t]
	\centering
	\captionsetup{justification=centering}
	\includegraphics[width=0.23\linewidth]{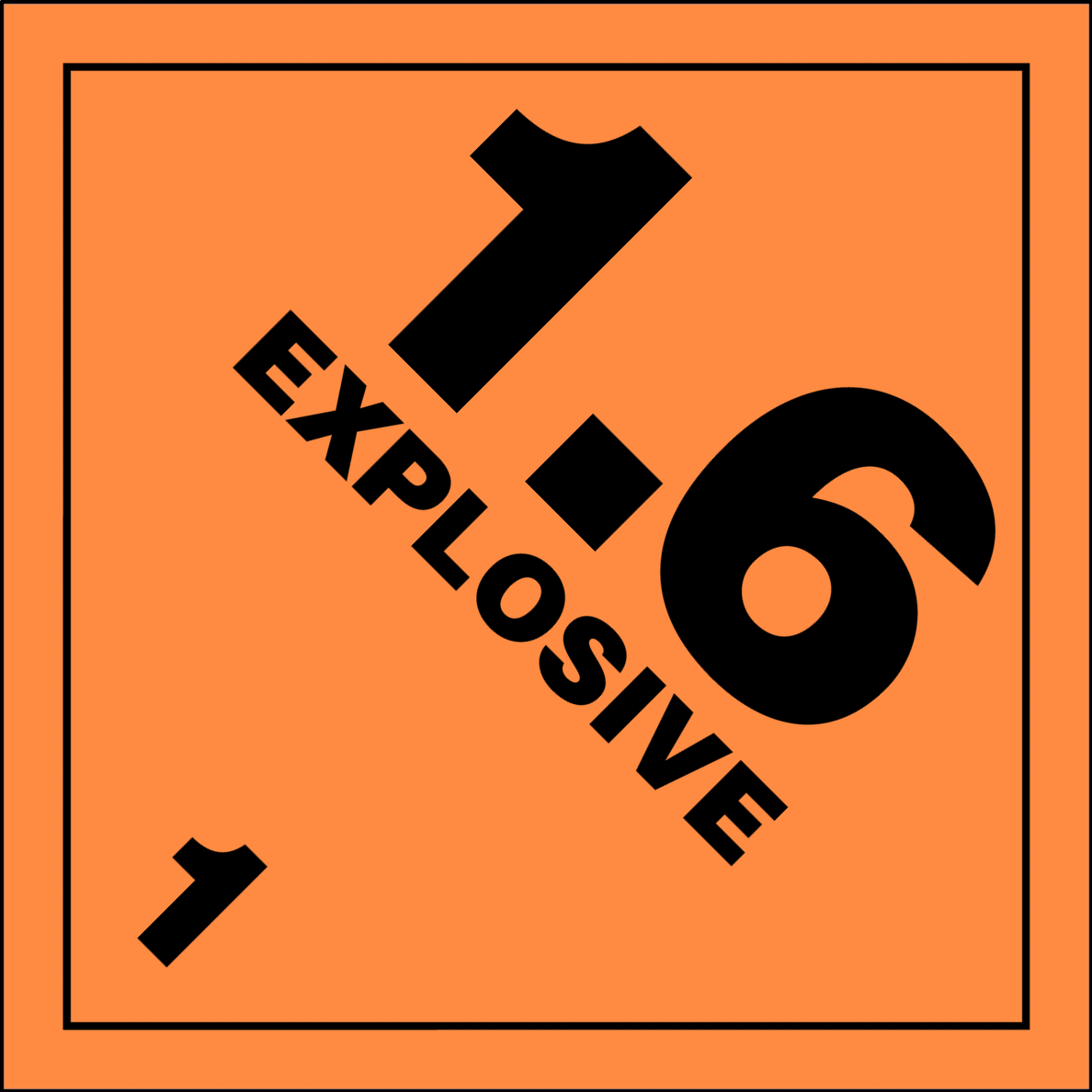}
	\label{fig:HazmatExample1}
	\includegraphics[width=0.23\linewidth]{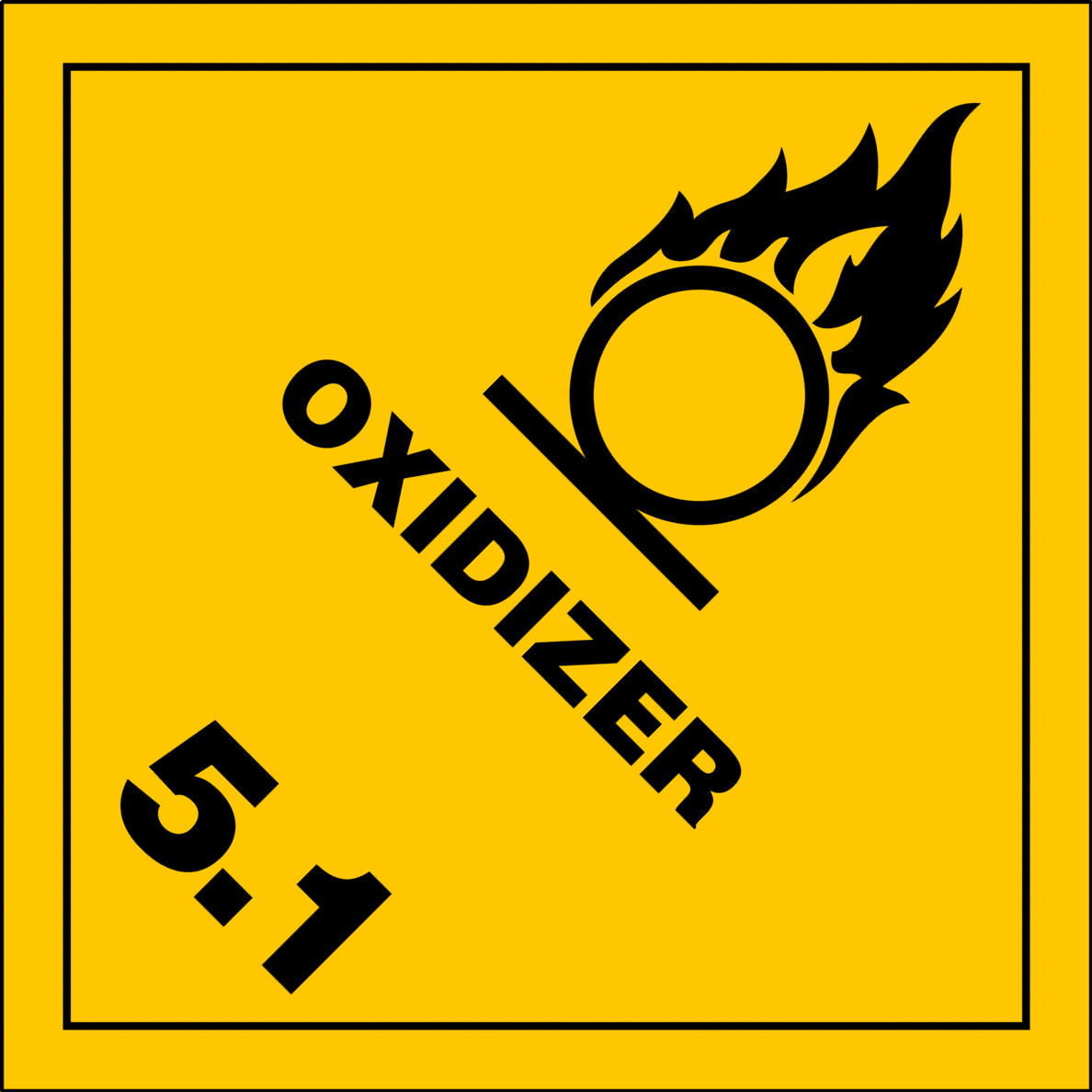}
	\label{fig:HazmatExample2}
	\includegraphics[width=0.23\linewidth]{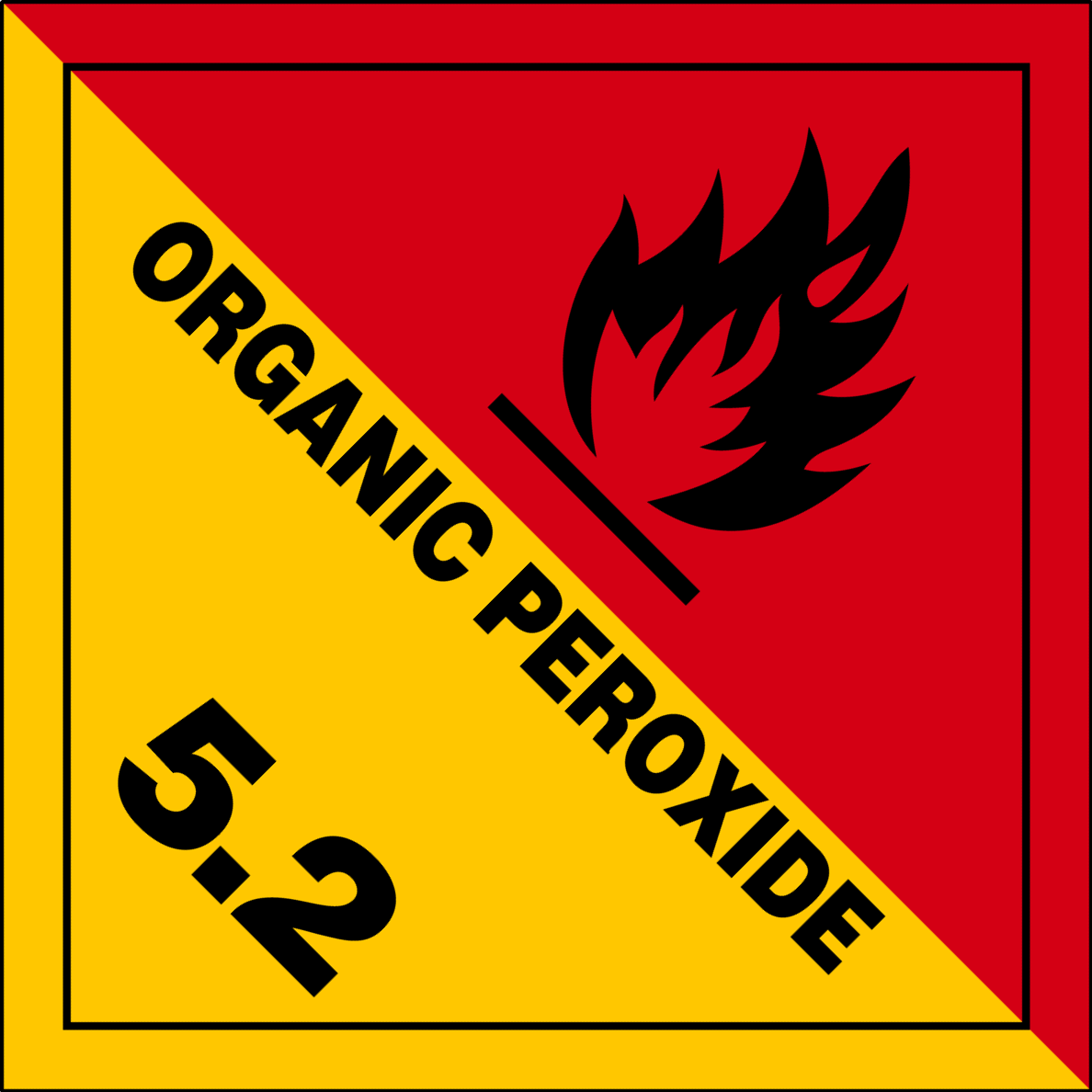}
	\label{fig:HazmatExample3}
	\includegraphics[width=0.23\linewidth]{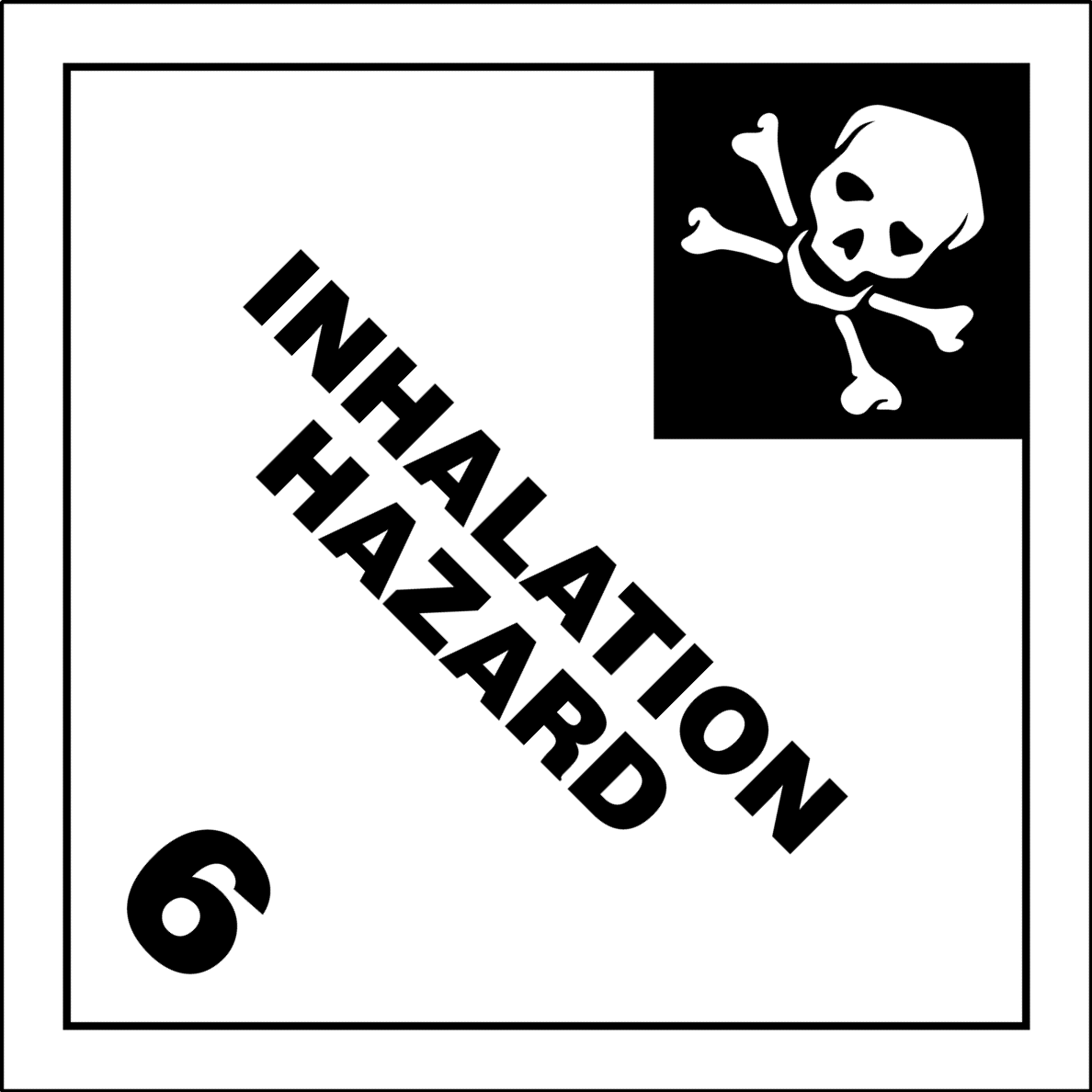}
	\label{fig:HazmatExample4}
	\caption{Hazmat Sign Reference Images. (GHS hazard pictograms)}
	\label{fig:HazmatExample}
\end{figure}

Affordable RGB-D sensors (such as Microsoft Kinect and Intel Realsense) are becoming more and more common in the modern robotics, due to their cheap price and portable size \cite{shan2019rgbd}. With the increasing attention on semantic understanding for mobile autonomous robots, we ask ourself the question: how might mobile robots take advantage of the depth information (RGB-D) for object detection in real-world robotics applications. In this work, we take hazmat sign (see Fig. \ref{fig:HazmatExample})  detection as an example task for using geometric image rectification to aid a CNN-based object detector.
\par
Hazmat sign detection has been studied by the robotics community for a long time. It is still challenging because of detection speed, illumination changes, background similarity, size variety, and inter-class variety. Due to the importance of hazmat sign detection for rescue robots, the RoboCup Rescue League competition has the task of detecting hazmat signs on the system inspection stage and in the exploration tasks\cite{Sheh2016robocup} \cite{sheh2011robocuprescue}. 
\par

Artificial Neural Networks (ANN) play a more and more important role in modern robotics \cite{sunderhauf2018limits} \cite{zhi2019learning}.
Visual object detection is the prime example for ANNs in robotics. Those methods can be categorized into two main approaches: Feature-based matching \cite{lowe2004sift} \cite{bay2006surf} \cite{rublee2011orb} and convolutional neural network (CNN) \cite{redmon2018yolov3} \cite{he2017maskrcnn} \cite{Zia2018DetectionRGBD}. Since the initial proposal of RCNN \cite{Girshick2014RCNN}, state-of-art CNN-based approaches \cite{redmon2018yolov3} \cite{he2017maskrcnn} have achieved impressive results on large-scale standardized datasets\cite{chen2015mscoco} \cite{kuznetsova2018openimage}. 
However, feeding images directly to CNN detector has two main drawbacks in terms of mobile robotics applications.
Firstly, to learn how to deal with distorted perspectives and multiple scales, data on different points of view is needed, which makes data collection and model training more time-consuming.
Secondly, even with a nice training set, there is no guarantee to learn perspective distortion because the network needs more layers for viewpoint angle estimation and perspective distortion, which can cause over-fitting in some cases.
\par
Moreover, previous research work has shown that it is hard for CNN detection frameworks to learn from raw RGB-D images automatically \cite{gupta2014learning}.  
Perspective distortion is introduced when the image is not captured in a canonical view of the object. Intuitively, parallel 3D lines are no longer parallel on the 2D image. This results in differences between images from different viewing angles. For planar objects, a homography matrix can be used to transfer between those images.

\par
In this work, we propose to utilize the depth information for rectifying RGB information with a  homography matrix. In short, a homography matrix is calculated from depth information to transform input RGB images to the canonical view. The CNN detector then takes rectified RGB images as input to perform detection. The final detection result on rectified RGB images is transformed back to the original images in the end.  
\par
The proposed method provides two key advantages. Firstly, the image rectification simplifies the problem for CNN detector.
Typical CNN detectors suffer from multi-scale input images and show noticeable weakness in bounding box regression accuracy. Image rectification can avoid the multi-scale problem to some extent. For the bounding box accuracy, as all images have been rectified to the canonical view, the problem gets slightly easier for CNN. Secondly, because the CNN detector only takes canonical-view images as input, the proposed method requires a smaller training set, which reduces the workload for practical deployment with mobile robotics.

\par
An efficient network should focus on the part of the problem which it is best at and which is most difficult for other methods. Therefore, skipping the learning of the perspective distortion is something for which we know exact models, such that a smaller and more efficient network can be used.




\par
In summary, our main contributions are:
\begin{itemize}
	\item We propose a feasible way of  combining geometric information with CNN detectors to improve detection performance. 
	\item Our approach shows good tolerance towards noise in depth images, with homography-based image rectification.
	
	\item We successfully reduce the number of images needed for training the CNN detector, because perspective distortion has been dealt with in advance. It is especially meaningful for practical usage in mobile robotics when facing a new environment or target object.  
	\item We release a new hazmat sign detection dataset. To the best of our knowledge, it is the first RGB-D hazmat sign detection dataset.
\end{itemize}

The remainder of this paper is organized as follows: 
Section \ref{sec:relatedWork} discusses related work while Section \ref{section:method} introduces the new detection framework with the homogrpahy-based rectification. Section \ref{section:dataset} presents the new RGB-D hazmat sign detection dataset. Experiments are evaluated in Section \ref{section:results} and conclusions are drawn in Section \ref{sec:conclusions}.

\section{Related Work}
\label{sec:relatedWork}
\label{Dataset::Related-Work}
\marky{
There are two main methods for image rectification. Kinahan et al. proposed a 3D projection method \cite{3Dreprojection}, and Chum et al. presented a 2D homography matrix transformation way \cite{homographies}.
}
In our approach, rectification via homography matrix is used, due to its robustness. Low-cost RGB-D cameras (like Intel Realsense) tend to have a quite noisy depth map. Through our experiments, we found that the pointcloud recovered from the depth map is too noisy for the other method. As it does not operate on 3D space, the homography method only requires the pointcloud to be accurate enough for plane extraction.

\par
Recent sign detection research is based on CNN methods, \marky{such as \cite{edlinger2019hazmat} proposed by Edlinger et al. and  \cite{Tnlist2016signdetection} presented by Zhu et al.} In our practical application, we found that detection results of deep learning methods, e.g. YOLOv3, are much better than classic vision approaches that extract keypoints. \marky{This concurs with the findings presented in \cite{edlinger2019hazmat}.} Thus, we use a deep learning method for sign detection in this task.


With the impressive performance of CNN networks, \marky{recent work mainly focuses on end-to-end CNN solutions like \cite{handa2016gvnn} proposed by Handa et al. and \cite{Zia2018DetectionRGBD} proposed by Zia et al.} 
\marky{Unfortunately, as Couprie et al. mentioned in \cite{couprie2013indoor}, the depth information with its high variance makes the learning process of CNN detectors even harder.} So feeding raw RGB-D images into CNN will not help much. 
\par
Some work instead utilizes depth information geometrically. Zhou et
al. presented a combination between CNN and geometry priors
in \cite{zhou2019monocap} by using CNN for image-based 2D part location estimates and assumes the geometry model for 3D pose reconstruction. \marky{Mousavian et al. utilize the geometric constraints on translation imposed by the 2D bounding box to recover a stable and accurate 3D object pose in \cite{mousavian20173d} }. \marky{Wang et al.  use geometric shape features to boost the performance of neural networks in \cite{wang2016differential}. Depth id used for proposal generation with contour detection in \cite{gupta2014learning} proposed by Gupta et al.} It also encodes the depth information before forwarding it to the  CNN detector.
\par
To the best of our knowledge, there is no concrete research on using image rectification for CNN-based object detection, 
which can take good advantage from the additional depth information. Nevertheless, there are some papers of image rectification for conventional feature matching.
\par
\marky{DoMonteLima et al. \cite{DoMonteLima2016RectificationRGBD} and Marcon et al. \cite{marcon20123d} use depth information to estimate surface normals to rectify the patch around the keypoints and use the rectified descriptors to improve the matching. The method proposed by DoMonteLima et al. in \cite{DoMonteLima2016RectificationRGBD} can be applied on both plane and non-plane cases in real-time.} However, both methods require feature detection before rectification, which does not fit to our task, because we need rectification pre-processing to improve the detection.
\par
Thus our task needs the rectification of the whole image or big planar sub-images, rather than that of patches. \marky{Eyjolfsdottir et al. aligned two images taken with a mobile phone by estimating their transformation in \cite{Eyjolfsdottir},} which requires accelerometer and gyroscope data from the phone. \marky{Gravity-aligned feature descriptors (GAFD) and gravity-rectified feature descriptors (GREFD) proposed by Kurzin et al. in  \cite{kurz2011gravity}} improve the matching performance of steep views by aligning the orientation of feature descriptors with the gravity. It also requires inertial sensors. 
\marky{The closest work to ours is \cite{wu20083d} presented by Wu et al..} They do the plane rectification before detecting features. However, they compute the local tangent plane's normal for each point on the surface, which has a high computation cost.

\section{Method}
\label{section:method}

\begin{figure}[t]
	\centering
	\includegraphics[width=0.6\linewidth]{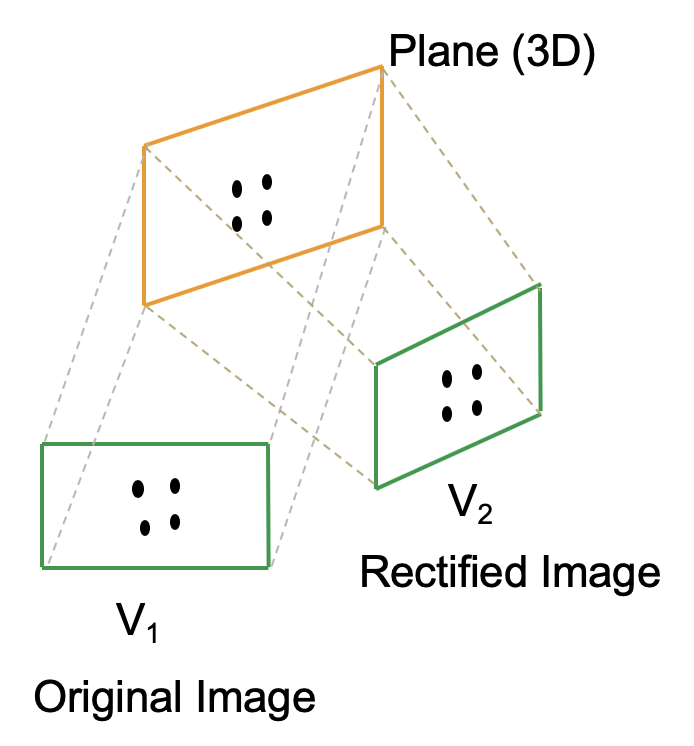}
	\caption{The proposed approach rectifies the image from the original viewpoint to a virtual viewpoint (canonical view).}
	\label{fig:rectification}
\end{figure}

%

\subsection{Overview Framework}
\label{subsec::overall_framework}

\par
The detection framework takes 3 inputs: 1) RGB images 2) point cloud 3) Camera intrinsic matrix. The Intel Realsense Driver provides the point cloud as output, so we use it as the input. This could be easily substituted with using depth images in the implementation. 
%
%

\subsection{Rectification Pipeline}
\label{subsec::rect_satrt}
\begin{figure*}[t]
	\centering
	\includegraphics[width=\textwidth]{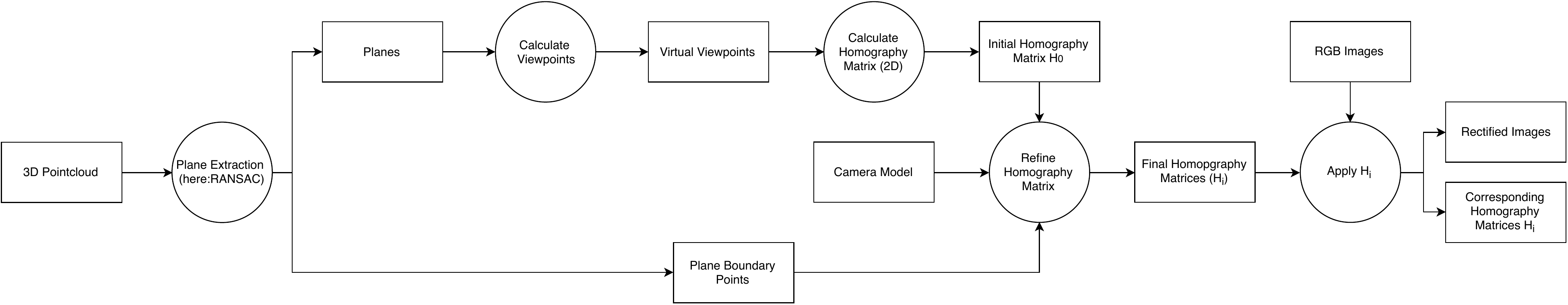}
	\caption{Overall Geometric Image Rectification Pipeline (Rectification Module)}
	\label{img::RectOverallPipeline}
\end{figure*}

The goal of the algorithm is to rectify the input images based on planes detected in the 3D data, as shown in Figure \ref{fig:rectification}. The overall pipeline contains the following stages and is shown in Figures \ref{fig::SystemOverallPipeline} and \ref{img::RectOverallPipeline}. The input image is rectified to multiple parallel viewpoints, so the rectification module needs to compute a set of homography matrices as the output. 
\begin{enumerate}
	\item Estimate plane segmentation from 3D point cloud;
	\item Calculate virtual canonical viewpoint;
	\item Compute the initial homography matrix for image rectification (rectify to the virtual canonical viewpoint);
	\item Refine the homography matrix by applying translation matrix (sliding through the image);
	\item Apply rectification matrices to get rectified images.
\end{enumerate}

\subsection{Plane Segmentation}
\label{sec:method_seg}
\subsubsection{Plane Estimation}
\par

RANSAC \cite{fischler1981ransac} is used to estimate plane parameters from the input point cloud. 
The usage of RANSAC shows a good robustness for plane estimation on noisy inputs, which then improve the robustness of the whole detection framework.


In this step, we obtain the major planes in the scene. For each plane, the  following are calculated (with respect to the original viewpoint, $v_1$):


\begin{itemize}
	\item the plane parameters $\pi$.
	$$\pi = (\mathbf{n^T}, d)$$
	where $\mathbf{n^T}$ is the normal of the plane, $||n|| = 1$, $d$ is the distance from viewpoint to plane.
	\item centriod point $\mathbf{P_c}$ 
	\item boundary points $\mathbf{X_{b\_i}}$ where $i \in \mathbf{Z^+}$
\end{itemize}


\subsubsection{Unique Normal}
\par
Each plane has two normals, in opposite directions. In order to calculate the new viewpoint, which has to be correctly aligned with the original viewpoint, it is necessary to use one unique normal for each plane. 
The unique normal is defined as the one not facing toward the origin. 
\begin{equation}
\mathbf{P}_c \cdot \mathbf{n^T} > 0
\end{equation}

Using the calculated normal we filter out the ground plane, as we don't expect any signs to detect there.

\subsection{Calculating the Virtual Canonical Viewpoint}
In order to transform images to the canonical view, a virtual canonical viewpoint needs to be calculated from the plane parameters. We set the virtual viewpoint ($v_2$) at a fixed distance away from the plane centroid. In our experiments, we set this to 1.2 meter to align with the training set, as the training set is collected with the distance of 0.8 and 1.2 meter. 

\par
The new viewpoint is calculated as: 

\begin{equation}
\mathbf{T}_2^1 = \begin{bmatrix}
\mathbf{R} & 0\\
0 &1 
\end{bmatrix}
*
\begin{bmatrix}
0 & \mathbf{t} \\
0 &1 
\end{bmatrix}
\end{equation}

\begin{equation}
\mathbf{t} = \mathbf{P}_c - d \hspace{0.25em} \mathbf{n^T}, \quad
\mathbf{R} = \begin{bmatrix}
0 & 0 & \mathbf{n}\\
\end{bmatrix}^T
\end{equation}

Where $T_2^1$ is a 4 * 4 matrix denoteing the position and orientation of $v_2$ in $v_1$. 
 $\mathbf{n^T}$ is the normal of the plane, $||n|| = 1$, $d$ is the distance from the viewpoint to the plane.
\begin{figure}[htbp]
	\centering
	\includegraphics[width=0.5\textwidth]{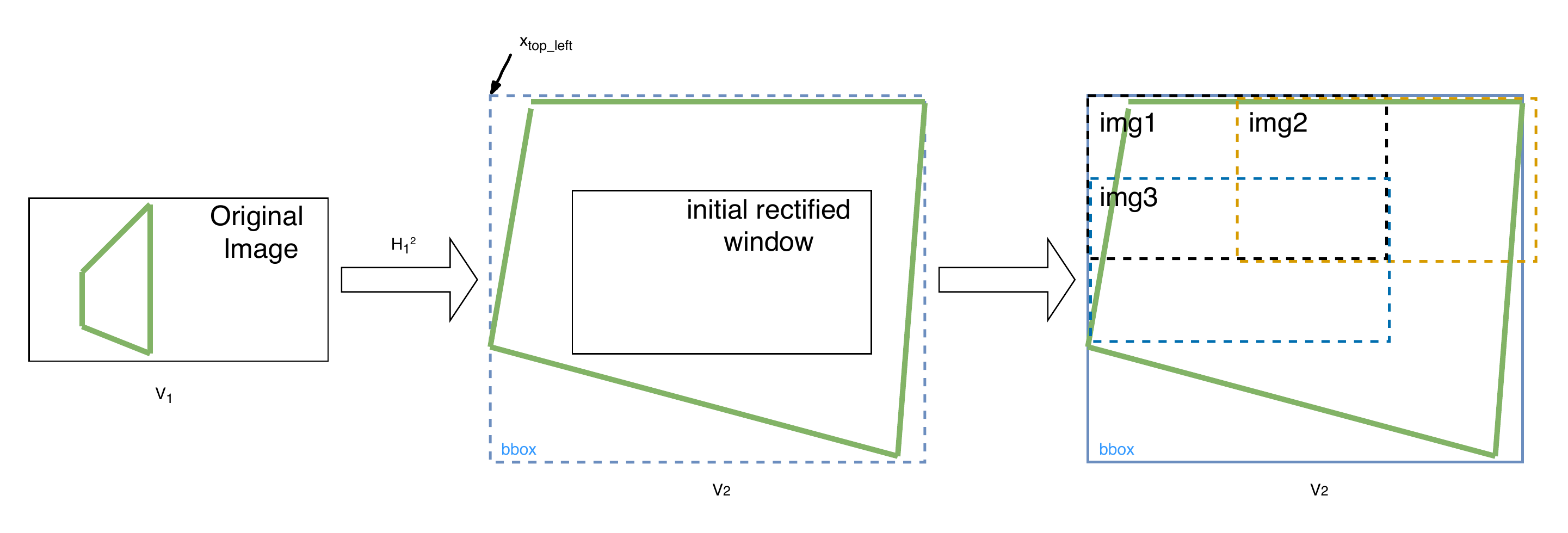}
	\caption{The middle image $V2$ shows the partial loss of the plane after applying $H_1^2$. Result image size is kept same as input. Green lines denote the border of a plane, solid rectangulars are image patches, the blue bounding box is the bounding box of the plane transformed to $V2$. The last image $V3$ is the visualization of system output matrices after applying the transformation. The resulting rectified images have a 50\% overlap with the neighbors. The result images are the rectangular boxes in black, orange and blue (img1, img2 and img3).}
	\label{fig::SlidingRefinement}
\end{figure}

\begin{figure*}[t]
	\centering
	\includegraphics[width=\textwidth]{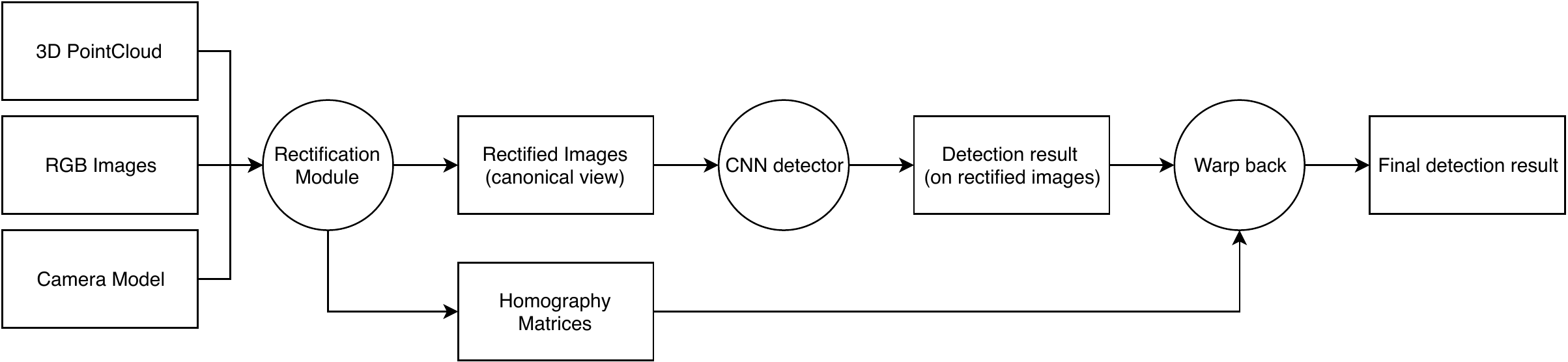}
	\caption{Proposed detection framework (testing pipeline)}
	\label{fig::SystemOverallPipeline}
\end{figure*}

\subsection{Calculate Homography Matrix for Virtual Viewpoint (2D)}
\par

Once the virtual viewpoint is calculated, it is easy to calculate an equivalent homography matrix that denotes the transformation between $v_1$ and $v_2$. Non-robust DLT (Direct Linear Transform) \cite{sutherland1974dlt} is used to compute the homography. 
\par
For homography calculation, four points are sampled from the 3D plane, denoted as $\mathbf{X}_i$. On both camera viewpoints (the original viewpoint and the new one), we project the 3D points to 2D images, denoted as $\mathbf{x}_1^{i}$ and $\mathbf{x}_2^{i}$. As shown in  Figure \ref{fig:rectification}.
\par
As the plane is an ideal infinite plane, and since the 2D image points are all obtained from re-projection, non-robust DLT (Direct Linear Transform) \cite{sutherland1974dlt} is used to compute the homography. 
\begin{equation}
I_2(x) = I_1(\mathbf{H} x)
\end{equation} where $x$ are homogenous coordinates.
\label{equation:homography}

\subsection{Refine Homography Matrix}
\label{sec::refine}
\label{subsec::rect_end}


At this point, we have a homography matrix which transforms side view images to the canonical view. 
However, there is still one minor issue: Some pixels go out of view, due to the FoV (field of view) of the camera. As shown in Figure \ref{fig::SlidingRefinement}, in some case, only part of the pixels are included in the resulting image. 

\subsubsection{Bounding box around the plane}
In our case, as the plane is a finite plane defined by boundary points, the first step is to calculate the tight bounding box around the reprojected plane in $v_2$, as is shown in blue in Figure \ref{fig::SlidingRefinement}. 
\begin{equation}
	\mathbf{x_b^2} = \mathbf{P} * \mathbf{T_1^2} * \mathbf{X_b}
\end{equation}
Where $\mathbf{X_b}$ denotes points around plane boundary, $P$ denotes projection matrix. 

From $\mathbf{x_b^2}$, we can easily calculate a tight bounding box by taking the  minimum and maximum. Specifically, we calculate the top-left corner $\mathbf{x_{top\_left}}$, the bounding box height $(H_{bbox})$ and the width $(W_{bbox})$.


\subsubsection{Refinement Algorithm}
This next step is to calculate the final rectification matrices which produce the final resulting images. As is shown in Equation \ref{equ::sliding}, we slide the camera window through the reprojected plane on $v_2$. Every two sliding images have 50\% of overlap either horizontally or vertically, so that hazmat signs around the border will present in the center of its sliding image. See Figure  \ref{fig::SlidingRefinement} for an example.

Denote $\mathbf{x_{top\_left}} = (x_{top\_left}, y_{top\_left})$ as the top-left corner of the plane bounding box on $v_2$, $H_{bbox}, W_{bbox}$ as the bounding box height and width. $H_{img}, W_{img}$ as height and width of the resulting image.

\begin{equation}
\begin{aligned}
\label{equ::sliding}
H &= \begin{bmatrix}
		1 & 0 & \frac{H_{img}}{2} * (i-1)\\
		0 & 1 & \frac{W_{img}}{2} * (j-1)\\
		0 & 0 & 1\\
\end{bmatrix}
* 
\begin{bmatrix}
			1 & 0 & -x_{top\_left}\\
			0 & 1 & -y_{top\_left}\\
			0 & 0 & 1
\end{bmatrix}\\
&= \begin{bmatrix}
1 & 0 & -x_{top\_left} + \frac{H_{img}}{2} * (i-1)\\
0 & 1 & -y_{top\_left} + \frac{W_{img}}{2} * (j-1)\\
0 & 0 & 1\\
\end{bmatrix}
\end{aligned}
\end{equation}

where i = 1 to 2 * $\lceil \frac{H_{bbox}}{H_{img}} \rceil - 1$ , $j = 1$ to 2 * $\lceil \frac{W_{bbox}}{W_{img}} \rceil - 1$

The final step is to apply the refined homography matrix to obtain the final rectified images. The warpping is done with bilinear interpolation. By doing that, object patches from non-canonical images are transformed into canonical view. As an additional benefit, the rectified images are all shown at the same distance, as we manually set it to a fixed distance (1.2 meters in this case). This avoids the multi-scale problem, which is challenging for CNN networks.

\section{Dataset}
\label{section:dataset}
\subsection{Static Scene Data}
\label{sec:static_data}
Currently, there are only very few publicly available hazmat sign detection datasets.  \cite{Mohamed2018hazmat} published their high-resolution RGB hazmat detection dataset. The dataset provided 600 high-resolution (5184*3456) RGB images containing hazmat signs from 5 different angles.
We provide a high-resolution RGB-D hazmat dataset, from an affordable RGB-D sensor (Intel Realsense RGB-D Camera), with $13$ labels, which can be found here\footnote{https://robotics.shanghaitech.edu.cn/datasets/MARS-Hazmat-RGBD}. It contains both $360$ RGB images and $360$ depth images with a resolution of $1280*720$. Ground truth label information of the RGB images is also provided. $130$ of the RGB and depth images contain only one of the $13$ types of hazmat label. Each of these images contains two backgrounds (plain and plywood) and five positions (top left, top right, center, bottom left, bottom right). For the rest of $230$ images, each image contains $13$ types of hazmat labels.   Nine different angles ($\pm75\degree$, $\pm60\degree$, $\pm45\degree$, $\pm30\degree$, $0\degree$)
with three distances ($1.25m$, $1.50m$, $1.75m$) are included in these images.

The dataset from \cite{Mohamed2018hazmat} was collected with a hand-held single-lens camera, containing only RGB images on static scenes. 
In contrast, our dataset contains color images as well as depth maps, with the additional depth information, which is able to provide  geometric information. 
Besides, our dataset includes 13  hazmat labels  while dataset \cite{Mohamed2018hazmat} only includes 8 labels. 

\section{Experiment and Results}
\label{section:results}
\subsection{Evaluation Metric}

\par
In our detection framework, we use a training set containing canonical-view images only, to train the CNN detector. It effectively reduces the size of the training set, thus lowering the difficulty for collecting a good training set for mobile robotics applications. 
In the testing stage, each image first goes through the rectification system to get rectified images in the canonical view, as is shown in Figure \ref{fig::SystemOverallPipeline}. Then the CNN detector is used for performing the actual detection. Finally, all detection results get warped back to the original images to get the final detection results as the output of the system. 

We use the MSCOCO \cite{chen2015mscoco} object detection evaluation matrix to evaluate the detection performance. The two main metrics are IoU (Intersection of Union) and mAP (mean Average Precision). We propose an extend NMS method to select final bounding boxes. Since each image is split into a series of images, we can convert the bounding boxes back to the origin image by utilizing the homography matrix we calculated before. 
For each splitted image we utilize the homography matrix to recover the final bounding boxes. The implementation of our method is provided here\footnote{https://github.com/STAR-Center/planar\_rect\_homography}.


\subsection{Experiments Setup}

We perform the experiments on our self-collected dataset, for there is no public available RGB-D hazmat detection dataset. To show that our approach can effectively reduce the difficulty for the CNN detector by dealing with perspective distortion in advance, only images from canonical views are used for training the network. The test sets include images from various angles (-75$^{\circ}$ to 75$^{\circ}$). 
\par
We use yolov3-tiny \cite{redmon2018yolov3} as the CNN detector. We choose yolov3-tiny because 1) We take it as an typical example of an off-the-shelf CNN-based detection network and 2) because it is small and fast enough for real-world deployment on mobile robots. The training time takes about one and a half hours with our computer (Intel Core i7-6700 CPU, GeForce GTX 1080, $8$ GiB Memory). 
We trained our model from scratch with a batch size of $32$, momentum $0.9$, subdivisions $2$,
burn$\_$in $2000$, max$\_$batches $8000$, learning$\_$rate $0.1$ and the learning$\_$rate will be multiplied by 0.1, when the number of batches is 3000, 4000, 5000, 6000, 7000.  
\par

\begin{table}[]
\vspace{0.8em}
	\caption{Result with / without geometry rectification. }
	\label{tab1} 
	\begin{tabular}{p{1cm}p{1cm}p{1cm}p{1.5cm}p{1.5cm}}
		\hline
		& mAP (IoU= 0.50)                & mAP(IoU= 0.75)                 & mAp(IoU= 0.50:0.05: 0.95)       & AR(IoU=  0.50:0.05: 0.95)       \\ \hline
		baseline & 0.236                          & 0.053                           & 0.088                           & 0.15                            \\ 
		our      & \textbf{0.53} & \textbf{0.193} & \textbf{0.246} & \textbf{0.351} \\ \hline
	\end{tabular}
\end{table}

\begin{table}[]
	\caption{Experiment result with / without geometry $~$ rectification in different angles. }
	\label{tab2}  
	\begin{tabular}{p{0.5cm}p{1cm}p{1cm}p{1cm}p{1.2cm}p{1.2cm}}
		\hline
		&          & mAP (IoU= 0.50)                 & mAP (IoU= 0.75)                  & mAP(IoU= 0.50:0.05: 0.95)        & AR(IoU= 0.50:0.05: 0.95)        \\ \hline
		\multirow{2}{*}{-75\degree} & baseline & 0.009                           & 0                               & 0.001                           & 0.004                           \\ \cline{2-6} 
		& our      & \textbf{0.132} & \textbf{0.006} & \textbf{0.034} & \textbf{0.051} \\ \hline
		\multirow{2}{*}{-60\degree} & baseline & 0.184                           & 0.016                           & 0.054                           & 0.077                           \\ \cline{2-6} 
		& our      & \textbf{0.477} & \textbf{0.055} & \textbf{0.158} & \textbf{0.216} \\ \hline
		\multirow{2}{*}{-45\degree} & baseline & 0.329                           & 0.099                           & 0.139                           & 0.195                           \\ \cline{2-6} 
		& our      & \textbf{0.645} & \textbf{0.274} & \textbf{0.303} & \textbf{0.401} \\ \hline
		\multirow{2}{*}{-30\degree} & baseline & 0.445                           & 0.149                           & 0.188                           & 0.272                           \\ \cline{2-6} 
		& our      & \textbf{0.679} & \textbf{0.365} & \textbf{0.362} & \textbf{0.456} \\ \hline
		\multirow{2}{*}{0\degree}   & baseline & 0.538                           & 0.222                           & 0.263                           & 0.357                           \\ \cline{2-6} 
		& our      & \textbf{0.665} & \textbf{0.386} & \textbf{0.375} & \textbf{0.492} \\ \hline
		\multirow{2}{*}{30\degree}  & baseline & 0.434                           & 0.09                            & 0.169                           & 0.247                           \\ \cline{2-6} 
		& our      & \textbf{0.632} & \textbf{0.367} & \textbf{0.364} & \textbf{0.455} \\ \hline
		\multirow{2}{*}{45\degree}  & baseline & 0.287                           & 0.041                           & 0.098                           & 0.145                           \\ \cline{2-6} 
		& our      & \textbf{0.663} & \textbf{0.361} & \textbf{0.364} & \textbf{0.46}  \\ \hline
		\multirow{2}{*}{60\degree}  & baseline & 0.116                           & 0.015                           & 0.029                           & 0.045                           \\ \cline{2-6} 
		& our      & \textbf{0.586} & \textbf{0.179} & \textbf{0.269} & \textbf{0.351} \\ \hline
		\multirow{2}{*}{75\degree}  & baseline & 0.026                           & 0                               & 0.004                           & 0.005                           \\ \cline{2-6} 
		& our      & \textbf{0.53}  & \textbf{0.114} & \textbf{0.222} & \textbf{0.281} \\ \hline
	\end{tabular}
\end{table}

\subsection{Rectification Parameters}
For the plane segmentation we assume that 90\% of the points of each frame are from planes. As a result, the RANSAC keeps extracting planes until less than 10\% of total points are in the remaining set.  
We set the number of maximum planes per image to 1, because it is known that the test set only contains one plane per test image. The distance from the virtual viewpoints to the  plane is set as 1.2m. Because the training dataset is collected with distances between 1m and 1.5m, it is reasonable to assume that the CNN detector will have better performance on 1.5m or 1m than others. The training dataset contains two main parts, images that contain one hazmat and images with 13 hazmats, in order to prevent overfitting to the background. The homography matrix can be calculated in a closed form solution by using planar homography, but we are using DLT to compute the homography matrix. In the future we plan to move to the closed form solution.

\subsection{Results}

We compare the detection performance with and without geometry rectification. The results are shown in Table \ref{tab1}. Baseline means without geometry rectification.
From Table \ref{tab1}, we can see that, after geometry rectification, the performance is much better than before. $mAP(IoU=0.50)$ increases nearly $20\%$ after geometry rectification while $mAP(IoU=
0.50:0.05:0.95)$ increases  $20.1\%$.

\begin{figure} [t]

	\begin{minipage}[t]{1\linewidth} %
		\centering   
		\includegraphics[width=0.95\linewidth]{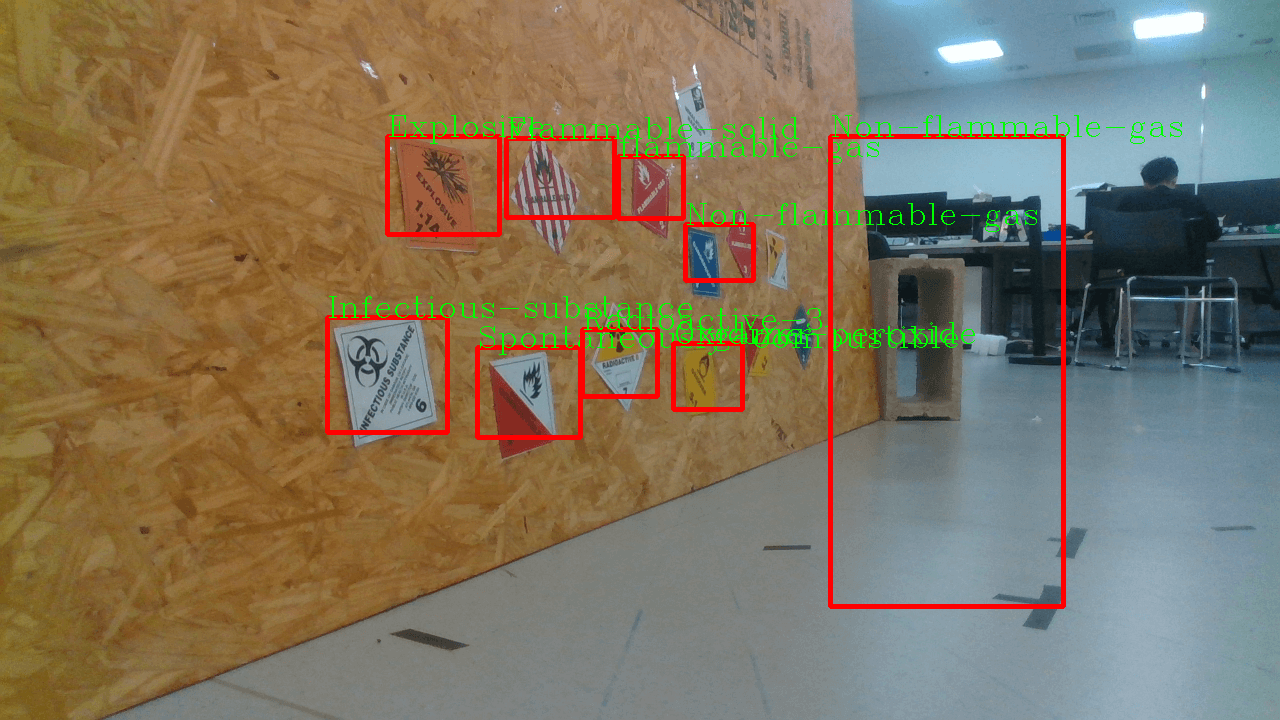}   \\\vspace{2mm}
		\includegraphics[width=0.95\linewidth]{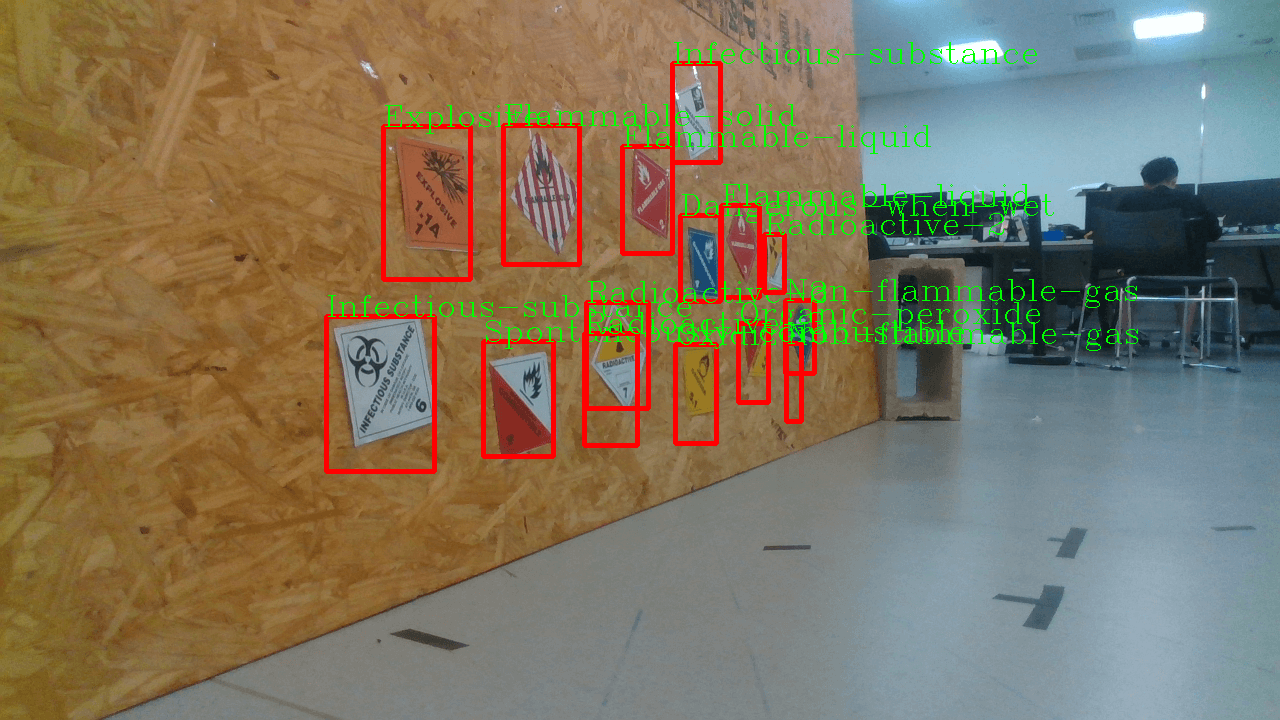}    

		\caption{Detection results at $60\degree$ on background one. Baseline on top, our approach at the bottom. }  
			\label{img::result60}   
	\end{minipage}%
\end{figure}

\begin{figure} [t] 
		\label{fig:result75}   
	\begin{minipage}[t]{1\linewidth} %
		\centering   
		\includegraphics[width=0.95\linewidth]{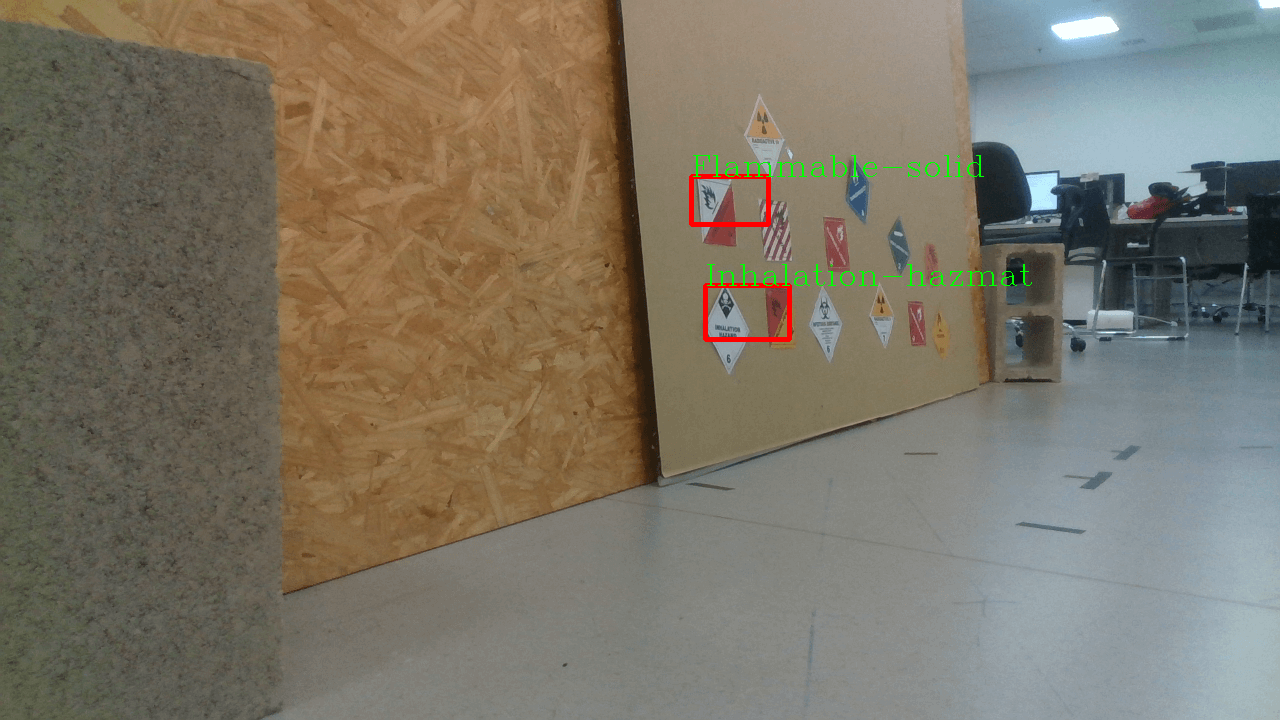} \\\vspace{2mm}

		\includegraphics[width=0.95\linewidth]{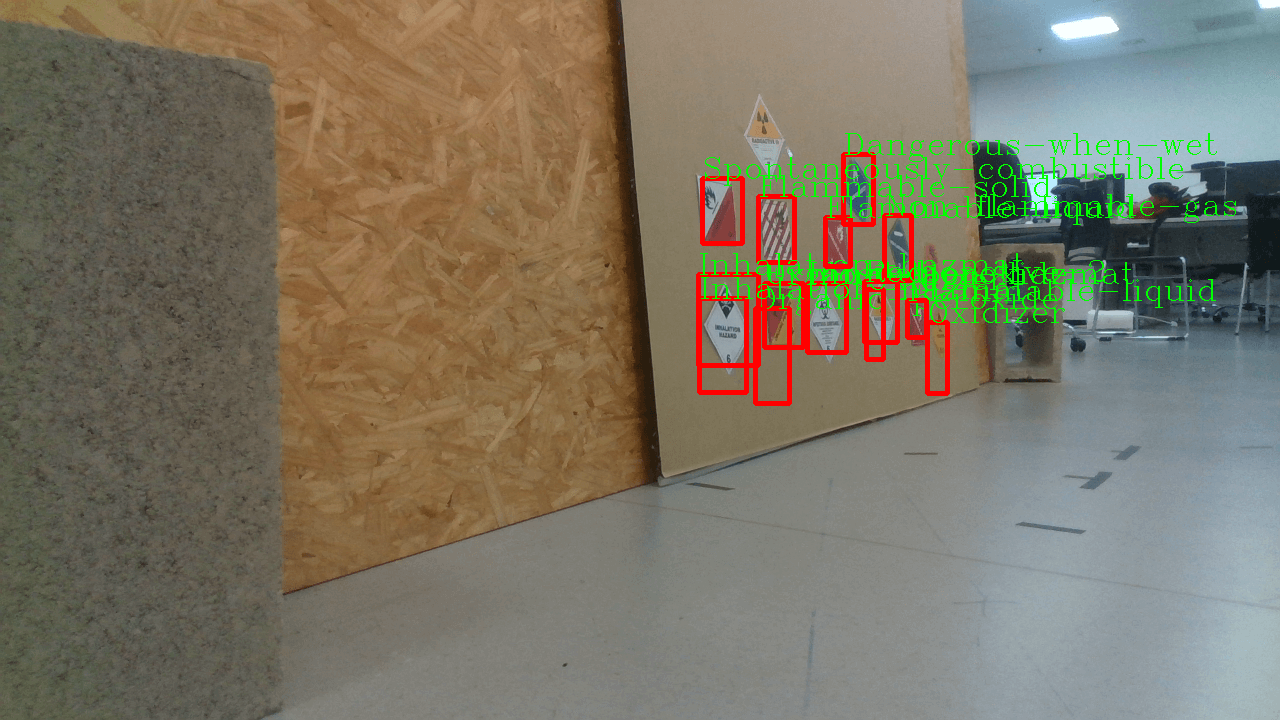}    

		\caption{Detection results at $75\degree$ on background two. Baseline on top, our approach at the bottom. } 
		\label{img::ab}    
	\end{minipage}%
\end{figure}

Table \ref{tab2} shows the performance of our approach at different angles. The test dataset contains  nine angles ($\pm75\degree, \pm60\degree, \pm45\degree, \pm30\degree, 0\degree$). From Table \ref{tab2}, we can see that, even at very large angles such as $\pm75\degree$, our approach can still detect some hazmat signs. In our approach the $mAP(IoU=0.5)$ is $0.132$ while without geometry rectification  $mAP(IoU=0.5)$ is $0.009$. At other angles such as $\pm60\degree,\pm45\degree, \pm30\degree $, according to the results shown in Table \ref{tab2}, our results are better than the results of the baseline approach. $mAP(IoU=0.5)$ increases $47\%$ at angle $60\degree$, which is a  huge improvement.

Besides perspective distortion, our proposed detection approach helps the CNN detector module to avoid dealing with multi-scale detection problems by explicitly rectifying the target patch to an ideal scale, as the distance from the plane to virtual viewpoints is fixed. Previous research has shown that the scale problem is challenging for CNN networks \cite{singh2018analysis}. As the result, in Table \ref{tab2}, we have better performance even at 0$\degree$, where there is no perspective distortion. For 0$\degree$ images, the mAP performance gets improved from 0.263 to 0.375.

The top images on Figure \ref{img::result60} and Figure \ref{img::ab} show examples of the detection results of baseline algorithm without geometry while the bottom images in  Figure \ref{img::result60} and Figure \ref{img::ab} show the detection results of our approach in the same images. We can see that on both Figure \ref{img::result60} and Figure \ref{img::ab} our approach performs much better than baseline approach. Our approach can detect more hazmat signs at very large angles. Also, the accuracy of bounding boxes is more precise in our approach.


\marky{Running single-threaded on a CPU, our algorithm needs
about 5.7 seconds per image, a value that should be improved,
since it is too slow to run live on a robot. This can be easily done by utilizing embedding GPU like Nvidia Jetson TX2 and using a multi-threaded implementation. 
We tested replacing RANSAC plane estimation with state-of-art fast plane extraction using agglomerative hierarchical clustering \cite{fastplane}, which lowers the inference time to around 1.5s. However, the robustness of the plane estimation is reduced.}








\section{CONCLUSIONS}
\label{sec:conclusions}
In this work, we showed a simple but effective way to combine geometric information with an off-the-shelf CNN-based detector. By doing image rectification explicitly in advance of the CNN detector, we take full advantage of available geometric information from RGB-D images to 1) reduce the time for training stage; 2) reduce the size of training set required; 3) improve the performance of overall detection system; and 4) produce more accurate detection result (tighter bounding box). This approach also features a high robustness towards noisy depth information input (noisy point cloud), as the depth is just used to estimate the plane parameters, where RANSAC is effective even with noisy input. 

For the mobile robotics application, especially hazmat sign detection in rescue robotics, our approach lowers the work required to create a nice training dataset, because fewer training images are needed. In the interest of reproducible science we provide the dataset used in the paper as well as our code to the public.







\IEEEtriggeratref{17}

\bibliographystyle{IEEEtran}
\bibliography{IEEEabrv,root}

\end{document}